\pdfoutput=1
\documentclass[11pt]{article}

\usepackage{EMNLP2023}

\usepackage{times}
\usepackage{latexsym}

\usepackage[T1]{fontenc}
\usepackage[utf8]{inputenc}

\usepackage{microtype}

\usepackage{inconsolata}

\usepackage{times,latexsym}
\usepackage{url}
\usepackage[T1]{fontenc}
\usepackage{amssymb}
\usepackage{amsmath}
\usepackage{cuted}
\usepackage{graphicx}
\usepackage{graphicx}
\usepackage{appendix}
\usepackage{esint}
\usepackage{bbold}
\usepackage{physics}
\usepackage{multirow}
\usepackage{multicol}
\usepackage{scrextend}
\usepackage{adjustbox}
\usepackage{algorithm}
\usepackage{algpseudocode}
\usepackage{xspace,mfirstuc,tabulary}

\newif\iftaclinstructions
\taclinstructionsfalse % AUTHORS: do NOT set this to true
\iftaclinstructions

\newcommand{\instr}
\fi

\title{Exploring the Limits of Fine-grained LLM-based Physics\\ Inference via Premise Removal Interventions}

\author{Jordan Meadows$^{1}$\thanks{* These authors contributed equally to this work.}, \; Tamsin James$^{2}$\footnotemark[1], \; Andr\'e Freitas$^{1,3}$ \\
$^1$University of Manchester, United Kingdom \\
$^2$University of Birmingham, United Kingdom \\
$^3$Idiap Research Institute, Switzerland \\
\texttt{jordan.meadows@postgrad.manchester.ac.uk} \\
\texttt{txj287@student.bham.ac.uk} \\
\texttt{andre.freitas@idiap.ch}}

\begin{document}
\maketitle
\begin{abstract}

Language models (LMs) can hallucinate when performing complex mathematical reasoning. Physics provides a rich domain for assessing their mathematical capabilities, where physical context requires that any symbolic manipulation satisfies complex semantics (\textit{e.g.,} units, tensorial order). In this work, we systematically remove crucial context from prompts to force instances where model inference may be algebraically coherent, yet unphysical. We assess LM capabilities in this domain using a curated dataset encompassing multiple notations and Physics subdomains. Further, we improve zero-shot scores using synthetic in-context examples, and demonstrate non-linear degradation of derivation quality with perturbation strength via the progressive omission of supporting premises. We find that the models' mathematical reasoning is not physics-informed in this setting, where physical context is predominantly ignored in favour of reverse-engineering solutions.

\end{abstract}

\section{Introduction}
\label{sec:intro}

%\begin{figure}
 %   \centering
 %   \includegraphics[width=1\linewidth]{paper6overview.PNG}
 %   \caption{An overview of the paper.}
 %   \label{fig:overview}
%\end{figure}

Language models demonstrate some level of mathematical ability~\cite{lewkowycz2022solving,liu2023evaluating,azerbayev2023llemma,pan2024quantum}. This reasoning modality requires controlled symbolic behaviour involving the repeated application of mathematical operations~\cite{valentino2023multi,meadows2023generating}, and LMs struggle to deliver this reliably~\cite{frieder2023mathematical,liu2024augmenting}. A particularly challenging mathematical domain is that of \textit{Physics}, where equation derivations serve as a rich environment within which the mathematical inference capabilities of LMs may be thoroughly examined, yet in contrast to other forms of mathematical reasoning~\cite{shakarian2023independent,yuan2023well,wang2023learning}, there are few examples of such efforts~\cite{lewkowycz2022solving}. While Mathematics proofs are closer to a logical argument over more abstract domain types, Physics derivations are centered around the integration of the abstraction of physical properties and laws into approximations and premises, where algebraic and calculus-related symbolic manipulations are performed to obtain novel equations, through a step-wise derivation, in close dialogue with empirical evidence.

An explanation for the lack of Physics-related approaches~\cite{luo2018automatic,wu2019toward,eivazi2022Physics,lewkowycz2022solving} is the fact that automating scientific discovery in mathematics has a long tradition in the context of automated theorem provers and proof assistants~\cite{jiang2022thor,lample2022hypertree}, and this relies on the translation of mathematical proofs into logical forms~\cite{szegedy2020promising,wuautoformalization2022}. However, much of Physics is incompatible with logical formalisation~\cite{kaliszyk2015formalizing,davis2019use,meadows2021similarity,yang2024leandojo,davis2024mathematics}. The flexibility of transformer-based models as soft reasoners~\cite{clark2020transformers} offers the opportunity to circumvent formalisation requirements, and develop models capable of detailed mathematical reasoning based on informally defined scientific knowledge. However, as with any form of mathematical reasoning, these models need to be built around models that support controlled, step-wise, symbolic inference.

One reason why current LMs fail in this regard, is because online resources used for training (such as Wikipedia and arXiv), and many mathematical datasets~\cite{mishra2022lila,hendrycks2021measuring,saxton2019analysing}, do not feature the required detail necessary for fine-grained reasoning. Physics derivations contain specific notations (\textit{e.g.,} Dirac notation) that are relied upon to build symbolically complex text spans, and complex operations (\textit{e.g.,} Laplace transform, Taylor expansion) form sophisticated dependencies between textual elements. Moreover, the derivations presented in papers and textbooks omit a significant number of steps, reinforcing difficulties in training models to perform more detailed calculations.

\begin{figure}[h!]
    \centering
    \includegraphics[width=0.8\linewidth]{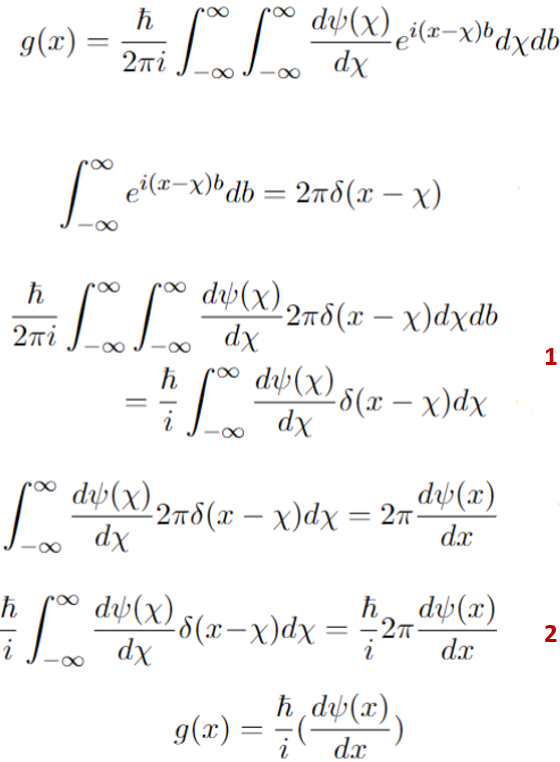}
    \caption{An incorrect derivation generated by few-shot GPT-4 that scores high ROUGE (81), BLEU (71), and GLEU (71). Erroneous equations are denoted in red.}
    \label{fig:gpt4}
\end{figure}

For instance, in Fig.~\ref{fig:gpt4}, GPT-4 fails a derivation (from the present data) related to the Uncertainty Principle in Quantum Mechanics. The first error arises from attempting to substitute the RHS of the second equation into the RHS of the first, but GPT erroneously keeps the $\int^{\infty}_{-\infty}db$ in the LHS. The second error involves an incorrect evaluation of an integral over a Dirac delta function (by a factor of $2\pi$), which is a critical result in Physics. Real-world research inherently involves equation manipulation which is out-of-distribution with respect to models' training data, either through a derivation's specific use of notation or its underlying reasoning. If state-of-the-art LMs fail at such basic manipulation, and if current evaluation metrics fail to account for such fine-grained errors~\cite{welleck2022naturalprover,meadows2023generating}, to what extent are leading methods appropriate for inference in mathematical domains~\cite{davis2024mathematics}?

This paper aims to explore these considerations, contributing with: \newline
\indent \textbf{(1.)} A manually curated step-wise granular dataset comprising 1200 derivation steps over 218 fine-grained Physics derivations at approximately graduate-level difficulty. Examples span a diverse set of subdomains including electromagnetism, classical, quantum, and statistical mechanics. The derivations are aligned to reference those on Wikipedia, but have been manually augmented to include finer steps, and are mapped to prompts containing premises and goal equations. \newline 
\indent \textbf{(2.)} Zero-shot and few-shot evaluation of GPT-4, GPT-3.5, and T5-related models on a \textit{Derivation Generation} task using common text generation metrics, paired with a manual evaluation, to highlight model reasoning limitations related to Physics. \newline \indent \textbf{(3.)} An exploration of models' out-of-distribution mathematical abilities centered on a controllable \textit{Premise Removal} intervention.

Through these contributions, we measure and qualify the ability of current LMs in performing out-of-distribution fine-grained multi-step mathematical reasoning, and aim to provide empirical foundations for developing transformer-based reasoners suitable for assisting discovery in Physics and related fields.

\begin{figure*}[h!]
    \centering
    \includegraphics[width=0.9\textwidth]{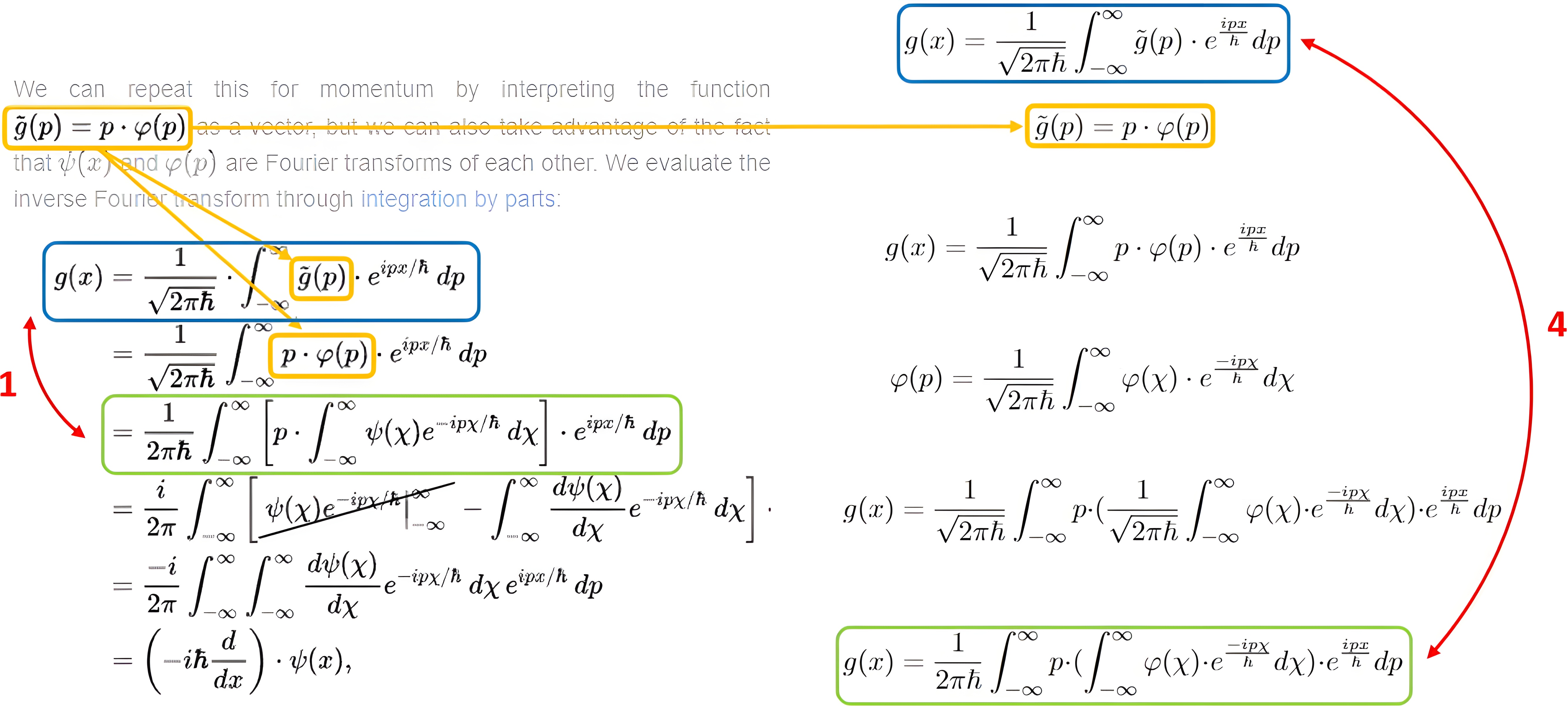}
    \caption{The difference between the Wikipedia proof (left) and our equational interpretation (right) of a reasoning chain related to the Uncertainty Principle in quantum mechanics. The (red) values represent the number of intermediate equations between equivalent equations in each representation, and highlights the detail gap.}
    \label{fig:proof_comp}
\end{figure*}

\section{Related work}

Our focus is on the generation of step-wise detailed derivations of Physics equations with LMs~\cite{brown2020language,ahmed2022few,song2022llm,ge2023openagi,hu2023llm,yang2023harnessing}. While we presently consider \textit{solely equations}, math generation exists in various forms, split between approaches that consider formal languages~\cite{first2023baldur,polu2022formal,jiang2021language,polu2020generative} and those considering informal mathematical natural language~\cite{ferreira2020natural,welleck2021naturalproofs,ferreira2022or,valentino2022textgraphs,meadows2023introduction}. Informal reasoning approaches generally involve code generation as input to symbolic solvers~\cite{heyueya2023solving,chen2022program,drori2022neural,mandlecha2022hybrid,hu2022enhancing,chen2021evaluating}, or directly generating math reasoning in natural language~\cite{lewkowycz2022solving,welleck2022naturalprover,chowdhery2022palm,lample2019deep}. Numerous approaches exist for evaluating the mathematical and symbolic capabilities and robustness of models~\cite{welleck2022symbolic,stolfo2022causal,meadows2023symbolic}, in contexts such as solving math word problems~\cite{roy2015reasoning,liang2022mwp,yao2023tree}. Various datasets exist containing mathematical reasoning~\cite{ferreira2020natural,hendrycks2021measuring,welleck2021naturalproofs,mishra2022lila}, and to a lesser extent, Physics~\cite{hendrycks2021measuring,lewkowycz2022solving,meadows2022physnlu,pan2024quantum}. We narrow our scope to isolate equational content from natural language descriptions, for the purpose of testing purely the equation manipulation capabilities of LMs in the Physics domain.

To summarise, we consider an equation Derivation Generation task, contribute a \textit{fine-grained} Physics dataset spanning multiple subdomains, perturb input prompts with a Premise Removal intervention, and explore model performance and degradation due to perturbations. We aim for this data to improve the detailed step-wise equation derivation capabilities of LMs, and use it to analyse their ability to perform Physics reasoning with complex equational forms (Fig.~\ref{fig:gpt4} and \ref{fig:proof_comp}). We later highlight the difference between coherent mathematical and physical reasoning (Fig.~\ref{tab:gpt4_analysis}), and discuss underlying generation degradation laws and how LMs perform mathematics in this context.

\section{Physics Dataset Construction}

%We describe the construction of resources that support the development and assessment of models capable of generating step-wise granular Physics derivations with a high level of mathematical detail. The target representation allows the assessment of the algebraic capabilities of models, as specific operations are isolated at a single step. 
%To reconstruct target representations models must perform sequential implicit operations to equations, such as integration and substitution, form implicit dependency graphs between linearised sequence elements during inference, while consistently using the correct notation and (LaTeX) syntax defined in the prompt.

To elicit the level of mathematical detail and coherence from models as described in Section \ref{sec:intro}, we randomly select a number of derivations from Wikipedia spanning Electromagnetism, Quantum, Classical and Statistical Mechanics, and expand each example until the required granularity is obtained (approximately one operation per step). We rely on the support of two annotators with adequate Physics expertise (Master's level).

Fig.~\ref{fig:proof_comp} gives an example of this rewriting and the departure from natural language. With respect to granularity differences, the single intermediate step (denoted by red) in the Wikipedia derivation on the left, is actually composed of multiple fine-grained steps naturally omitted for the sake of succinct communication online. Experienced physicists may indeed skip these steps within their own workings, but there is no guarantee that LMs can reliably perform them without error~\cite{welleck2022naturalprover,frieder2023mathematical,meadows2023generating,liu2024augmenting, quan2024verificationrefinementnaturallanguage}. On the right-hand side, this single step is expanded into four finer-grained steps that \textit{improve explainability}. The first and second equations in the expanded derivation are premises extracted from the text. The third equation is formed by substituting the second premise into the first, and the fourth equation is not explicitly written anywhere -- it is the Fourier transform of $\varphi(\chi)$ described non-mathematically within the initial description now included for completeness. The remaining equations are obtained through substitution.

We algorithmically describe this expansion and annotation process for converting online derivations into examples rendered in Appendices~\ref{app:electromagnetism}-\ref{app:other}, in Alg.~\ref{alg:annotation}. This protocol was applied by each annotator within a double swap, review and refine setting. Initial derivations were split between each annotator, and after initial annotation, the datasets were swapped for review and changes were tracked. A second swap was performed for accepting the proposed changes.

The inclusion of the (implicit) Fourier transform and the (explicit) premise in Fig.~\ref{fig:proof_comp}, give an example of how a combined modality of natural language and equations is mapped into a single equational modality. The expansion of one step into four, describes our push towards the explainable reasoning we wish to elicit from models.

High-quality~\cite{villalobos2022run} data of this kind is necessary to bridge an incompleteness gap between the communicative reasoning available online (used to train/evaluate models) and the reality of Physics calculations~\cite{pan2024quantum,akrobotu2022qubo,meadows2021similarity,mann2018manipulating,hopfield1958theory}. Establishing that models can robustly perform such reasoning is a prerequisite enabling their utility and application in theoretical discovery.

\begin{algorithm}[h!]
\caption{Derivation Annotation}
\begin{algorithmic}[1]
        \State \texttt{Define premises} from a given initial derivation.
        \State \texttt{Derive intermediate equations} between initial equations.
        \State \texttt{Re-write derivation} asserting one operation per step (approximately).
        \State \texttt{Write derivation in LaTeX} as a sequence of equations.
        \State \texttt{Annotate} ``\%PREM'' within equation environments of premises.
        \State \texttt{Re-organise LaTeX derivation} into self-contained 4-9 step sub-derivations.
        \State \texttt{Re-annotate} new sub-derivation premises with ``\%PREM''.
        \State \texttt{Output multiple derivations} in LaTeX per initial derivation.
\end{algorithmic}
\label{alg:annotation}
\end{algorithm}

\paragraph{Data Analysis.} We expand 60 derivations from Wikipedia examples following the discussed protocol, and extract 218 shorter examples resembling the right-hand side of Fig.~\ref{fig:proof_comp}. These are split into three categories: \textbf{Electromagnetism} (containing vector calculus), \textbf{Quantum Mechanics} (containing Dirac notation and commutation relations), and \textbf{Other Physics} (containing results from \textit{classical} and \textit{statistical mechanics}). Tab.~\ref{tab:sizes} describes the number of examples within each subdomain, and example derivations are rendered in Appendices~\ref{app:electromagnetism} (electromagnetism), \ref{app:quantum} (quantum), and \ref{app:other} (other).

{\renewcommand{\arraystretch}{1.1}%
\begin{table}[h!]
	\centering
	\scalebox{1}{
		\begin{tabular}{@{}c|c @{}}
			\textbf{Field of Physics} & \textbf{\# Derivations}\\
			\hline
            Electromagnetism & 82 \\
            Quantum Mechanics & 98 \\
            Other Physics & 38 \\
            All fields & 218 \\
		\end{tabular}
		}
	\caption{Number of derivations in the dataset by field of Physics. The \textit{Other} category corresponds to results from classical and statistical mechanics.}
	\label{tab:sizes}
\end{table}}

\begin{figure}[h!]
    \centering
    \includegraphics[width=0.85\linewidth]{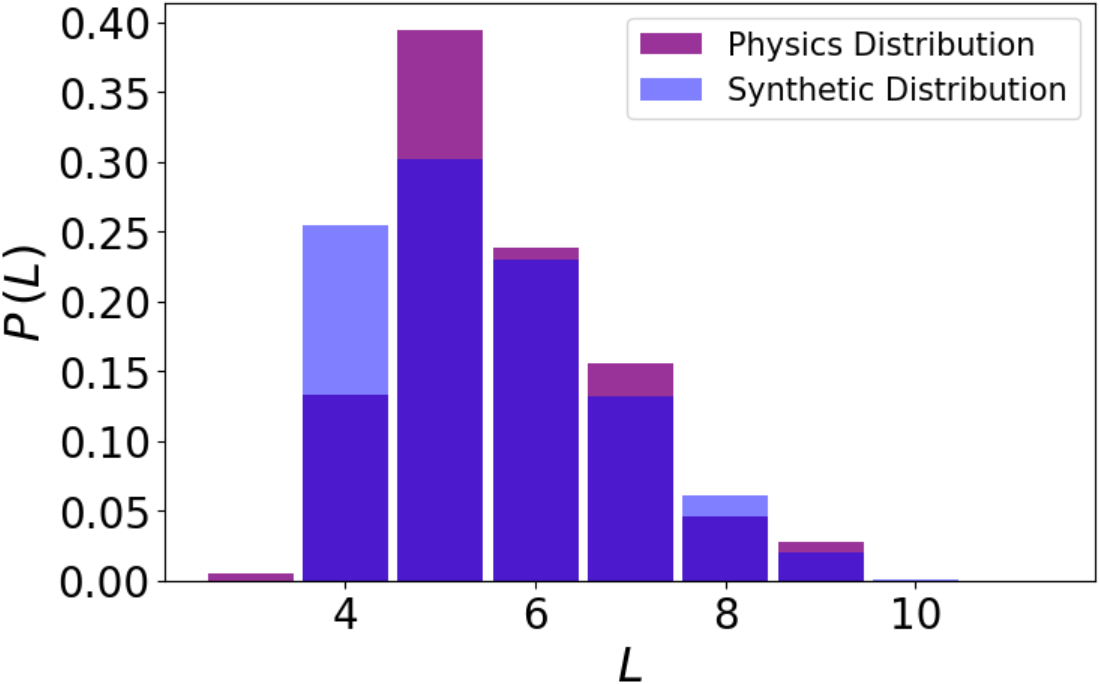}
    \caption{$P(L)$ is the probability that a given derivation contains $L$ equations.}
    \label{fig:length_distribution}
\end{figure}

Smaller models, such as BERT~\cite{devlin2018bert} and T5~\cite{raffel2020exploring}, are limited to input sequences of up to 512 tokens. This corresponds to a LaTeX derivation comprising between 4-9 equations. The original expanded derivations are segmented into shorter examples to accommodate these limitations. The relevant length distributions are given in Fig.~\ref{fig:length_distribution} alongside that of the synthetic dataset used to fine-tune MathT5 in related work~\cite{meadows2023generating}, and to provide in-context examples for few-shot prompts in these experiments. Other similarities with the synthetic dataset include an overlap of 155 symbols and a similar step granularity.

Otherwise, the Physics derivations contain significantly more symbol combinations, including limits of integration and entirely separate notation. In particular, this includes \textit{Dirac notation} that is commonplace in Quantum Mechanics (such as $\int \braket{x}{\Psi}^{\dagger}x'\delta(x-x') \braket{x'}{\Psi} dx' = \braket{x}{\Psi}^{\dagger}x \braket{x}{\Psi}$, from \ref{app:quantum}.3), and \textit{vector calculus} involving the div, grad, and curl operators which are commonplace in Electromagnetism (such as $\nabla \cdot (\phi\nabla\phi) = (\nabla\phi)^2 + \phi\nabla^2\phi$, from \ref{app:electromagnetism}.2).

\begin{figure}[h!]
    \centering
    \includegraphics[width=0.85\linewidth]{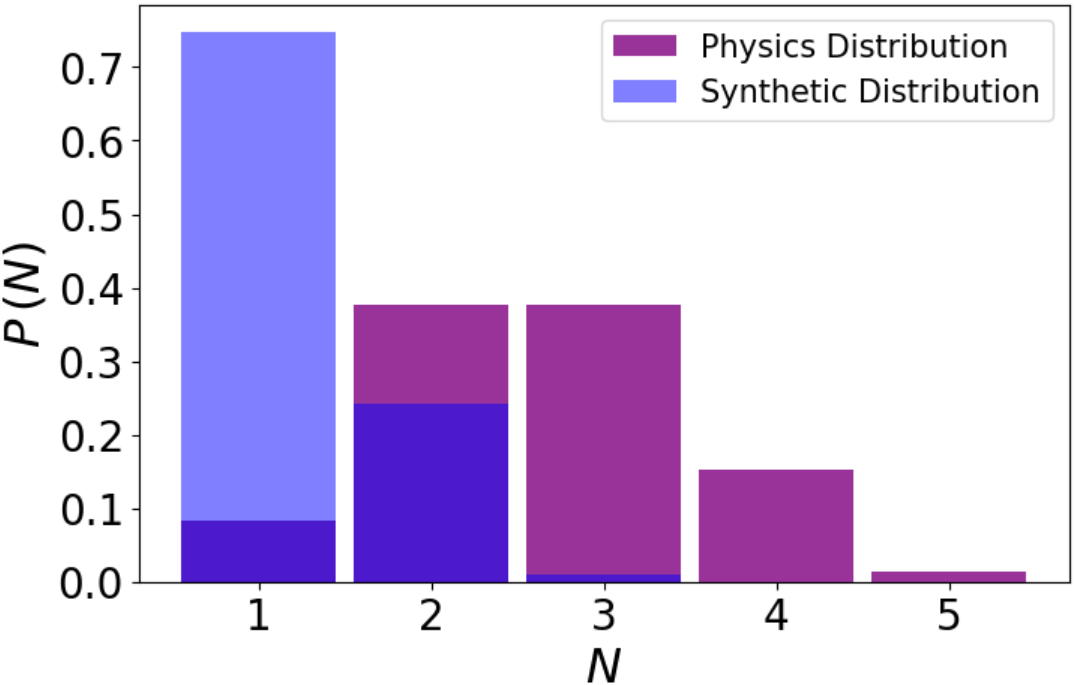}
    \caption{$P(N)$ is the probability that a given derivation contains $N$ premise equations.}
    \label{fig:premise_distribution}
\end{figure}

\textit{Premises} are another crucial element of derivations. These are axiomatic equations deemed necessary for deriving a goal equation. The distribution of the number of premises per derivation is given in Fig.~\ref{fig:premise_distribution}. A significant proportion of steps involves simply writing the premises and goal equation in the correct order with correct syntax (a non-trivial task for LMs~\cite{chen2024premise,meadows2023generating}), or substituting expressions.

A final noteworthy property of the dataset is its compositionality towards longer derivations. Premises are separated from non-premises, and goal equations from some examples are premises in another, meaning that sequences may be chained together to form much longer examples of up to 20 steps. The dataset itself contains both prompts and target derivations. Output text takes the form of LaTeX equations conjoined by an ``and'' token, which provides a minimalistic template for defining equation boundaries. The dataset is available online\footnote{\url{https://github.com/jmeadows17/transformers-for-physics}}, and includes lists of derivation-specific premises and references to original Wikipedia names alongside each example.

\section{Derivation Generation and Generalisation Capabilities}

The experimental analysis occurs in two parts: evaluation on an in-distribution test set in the setting of a Derivation Generation task, followed by an out-of-distribution evaluation facilitated by the progressive removal of premises from in-distribution prompts. Perturbed prompts are out-of-distribution with respect to either a model's training data or in-context examples.

\subsection{The Derivation Generation task}
\label{sec:derivation_generation}

Given a goal equation $G$ and premises $\mathcal{P}$ arranged within some prompt template $t(\mathcal{P}, G)$, a given model $\mathcal{M}$ must generate a sequence of equations $\mathcal{\hat{D}}$ which represents a reasonable derivation of $G$. A derivation is generated through $\mathcal{M}:t(\mathcal{P}, G) \mapsto \mathcal{\hat{D}}$, which is then compared to an idealised ground truth $\mathcal{D}^*$. Some idealised metric $M^*$ scores the generated derivation through $M^*:(\mathcal{D}^*,\hat{\mathcal{D}}) \mapsto \mathcal{S}$. Assuming a suitable prompt $t$, we generally aim to optimise the following problem to find the best model according to the given metric, through:
\begin{equation*}
    \mathcal{M}^* = \underset{\mathcal{M}}{\mathrm{argmax}}; M^*\big(\mathcal{D}^*, \mathcal{M}:t(\mathcal{P}, G) \mapsto \mathcal{\hat{D}}\big).
\end{equation*}

\noindent However, we do not have access to ideal derivations $\mathcal{D}^*$ corresponding to templates $t(\mathcal{P}, G)$, as many derivations may reasonably derive $G$, yet may differ from $\mathcal{D}^*$. We also do not have access to ideal metric $M^*$ suitable for accurately scoring individual $\hat{\mathcal{D}}$. Instead, we manually produce coherent ground truths $\tilde{\mathcal{D}^*}$ and \textit{assume} that the quality of model derivations is reflected monotonically (on average) with scores obtained from the text generation metrics. We then conventionally determine $\mathcal{M}^*$ through
\begin{equation*}
\scalebox{0.85}{
    $\mathcal{M}^* = \underset{\mathcal{M}}{\mathrm{argmax}}; \frac{1}{N}\sum_{i=1}^N M\big(\tilde{\mathcal{D}^*}_i, \mathcal{M}:t(\mathcal{P}_i, G_i) \mapsto \mathcal{\hat{D}}_i\big),$
}
\end{equation*}

\noindent where $N$ is the number of ground truth derivations. In this work, we consider $M$ as a reference-based generation metric (\textit{e.g.,} ROUGE) used to evaluate derivations at scale, but we contrast this with a reference-free human evaluation of derivations. An example prompt $t(\mathcal{P}, G)$ is given below.

\vspace{10pt}
\begin{addmargin}[1em]{1em}% 1em left, 2em right
\textit{Given \;\; $q{(a)} = e^{a}$ \newline \newline and  \;\; $G{(a)} = - e^{a} + \frac{d}{d a} q{(a)}$, \newline \newline
then obtain \;\; $e^{G{(a)}} = 1$}
\end{addmargin}
\vspace{10pt}

\paragraph{Prompting LMs.} The specific details for zero-shot and few-shot prompts are described in Appendix~\ref{app:models_and_metrics}, but we give a brief summary here. For zero-shot prompts, a simple task description is prefixed to the above template which emphasises the equational focus of the output template. For few-shot prompts, a total of 5 in-context examples are selected from the synthetic training set (used to train MathT5) following \citet{meadows2023generating}. We note an inherent similarity with chain-of-thought prompting~\cite{wei2023chainofthought} due to the nature of the task, but prompts deviate from this due to the exclusion of natural language. 

\subsection{Controlled Premise Removal}
\label{sec:perturbation}
The goal of premise removal is to systematically remove crucial mathematical and physical context from prompts, such that a given model must either recall this missing context \textit{and} use it appropriately, or derive the goal equation via an alternative route. If a model understands the underlying reasoning, then the removal of premises should not incur derivation errors. If errors do occur as a function of the number of premises removed, we can study how certain reasoning capabilities degrade as the strength of the perturbation increases.

The prompt template accommodates the removal of premises without introducing uncontrolled perturbations to other mathematical terms or natural language. This relatively pure intervention on the input space~\cite{stolfo2022causal,pearl2009causal} introduces a secondary \textit{premise selection} problem~\cite{alama2014premise,wang2017premise,ferreira2020natural,valentino-etal-2022-textgraphs,meadows2023introduction} in tandem with the main task. If successful Derivation Generation involves the application of mathematical operations to equations during inference, such as algebraic manipulation and calculus, then Premise Removal introduces the requirement that models must either generate a missing premise, or find an alternative derivation route. We can increase the severity of the distribution shift (Fig.~\ref{fig:premise_distribution}) by progressively removing premises. 

More formally, if the prompt $t(\mathcal{P}, G)$ is a template containing (ordered) premises $p_i \in \mathcal{P}$, and the goal equation $G$, through string concatenation ($+$) we can define this perturbation as 
\begin{equation}
\label{sec:equation}
    \resizebox{0.485\textwidth}{!}{$t(\mathcal{P}, G, S; \alpha, \beta, \gamma) = \alpha + p_1 + \sum_{i=2}^{|\mathcal{P}| - S - 1}(\beta + p_i) + \gamma + G$}
\end{equation} 

\noindent where $\alpha, \beta, \gamma$ are natural language sequences held constant with respect to any $p_i$ or $G$. Importantly, $S$ \textit{directly controls the perturbation strength}. $S$ premises are removed from the prompt (reverse chronologically) such that if $S = 0$ we recover the original in-distribution prompt. In this work we consider $S \in \{0, 1, 2\}$, and are hence restricted to derivations containing $|\mathcal{P}| \geq 3$ (about half of the data, Fig.~\ref{fig:premise_distribution}), as at least one premise is required by the prompt template.

To briefly demonstrate, the example prompt in Section \ref{sec:derivation_generation} corresponds to $(|\mathcal{P}|, S) = (2, 0)$, whereas $(|\mathcal{P}|, S) = (2, 1)$ corresponds to the prompt below (\textit{i.e.,} $\alpha + p_1 + \gamma + G$). Both would be excluded from this analysis as $|\mathcal{P}| = 2$.

\vspace{10pt}
\begin{addmargin}[1em]{1em}% 1em left, 2em right
\textit{Given \;\; $q{(a)} = e^{a}$ \newline \newline
then obtain \;\; $e^{G{(a)}} = 1$}
\end{addmargin}
\vspace{10pt}

\section{Evaluation}

We describe details of models and metrics in Appendix~\ref{app:models_and_metrics}. Here we report key results from the Derivation Generation experiments, beginning with an evaluation of unperturbed prompts (Section \ref{sec:unperturbed_results}) followed by an exploration of performance degradation due to Premise Removal perturbations (Section \ref{sec:premise_removal}). 

\subsection{Derivation Generation}
\label{sec:unperturbed_results}

{\renewcommand{\arraystretch}{1}%
\begin{table}[h!]
\centering
\scalebox{0.85}{
\begin{tabular}{@{}l|c|c|c|c @{}}
& \textbf{ROUGE} & \textbf{BLEU} & \textbf{GLEU}\\
\hline
T5-base & 13.6 & 1.8 & 7.2 \\
T5-large & 9.5 & 1.0 & 5.0 \\
FLAN-T5-base & 9.2 & 2.7 & 6.4\\
FLAN-T5-large & 7.0 & 0.6 & 3.5\\
MathT5-base & 70.6 & 58.6 & 60.7\\
MathT5-large & 69.6 & 60.2 & 62.1 \\
GPT-3.5 \textit{(ZS)} & 56.8 & 48.3 & 51.3 \\
GPT-3.5 & 77.7 & 67.3 & 70.3 \\
GPT-4 \textit{(ZS)} & 77.1 & 60.3 & 66.4 \\
GPT-4 & \textbf{84.3} & \textbf{78.0} & \textbf{79.1}
\end{tabular}
}
\caption{Evaluation results for derivation generation with the Physics dataset. \textit{(ZS)} refers to zero-shot performance. Otherwise, all T5 results are zero-shot, and GPT results are few-shot.}
\label{tab:main_scores}
\end{table}}

\noindent \textbf{Synthetic in-context examples improve inference.} The few-shot approach used to prompt the GPT models does not use Physics derivations as in-context examples, and therefore \textit{does not learn any biases present in the Physics data}. Instead, synthetic derivations are included in prompts that demonstrate the required granularity between steps, while helping to force model output into the required template. Few-shot learning over general mathematical text significantly improves performance over zero-shot prompting, and notably, few-shot GPT-3.5 outperforms zero-shot GPT-4 across all metrics in Tab.~\ref{tab:main_scores}.

%\textbf{MathT5 outperforms zero-shot GPT-3.5 and all T5 models.} The MathT5 models both perform similarly, and are outperformed only by GPT-4 in zero-shot performance. The advantage over GPT-3.5 is surprising, considering the evaluation considers significantly out-of-distribution derivations (\textit{e.g.,} Fig.~\ref{fig:length_distribution} and \ref{fig:premise_distribution}), and the fact that GPT-3.5 contains three orders of magnitude more parameters than MathT5-base. Despite known mathematical limitations of GPT-3.5~\cite{frieder2023mathematical}, this discrepancy likely comes from two sources: GPT-3.5's inability to constrain the output derivations to the required format, and secondarily that MathT5 is fine-tuned on fine-grained synthetic derivations from the original T5/FLAN-T5 weights.

{\renewcommand{\arraystretch}{1}%
\begin{table}[h!]
\centering
\scalebox{0.8}{
\begin{tabular}{@{}l|c|c|c|c|c @{}}
& \textbf{Field} & \textbf{ROUGE} & \textbf{BLEU} & \textbf{GLEU} \\
\hline
\multirow{3}{*}{MathT5-base} & \textit{EM} & 75.9 & 64.8 & 66.3\\
& \textit{QM} & 68.4 & 55.7 & 58.2 \\
& \textit{Other} & 63.2 & 50.6 & 52.9\\
\hline
\multirow{3}{*}{MathT5-large} & \textit{EM} & 75.1 & 67.9 & 69.1\\
& \textit{QM} & 67.9 & 57.3 & 59.2\\
& \textit{Other} & 60.0 & 48.6 & 52.3\\
\hline
\multirow{3}{*}{GPT-3.5} & \textit{EM} & 82.7 & 71.1 & 75.3\\
& \textit{QM} & 76.7 & 66.7 & 69.1\\
& \textit{Other} & 67.6 & 59.1 & 61.1\\
\hline
\multirow{3}{*}{GPT-4} & \textit{EM} & 86.4 & 81.0 & 81.8\\
& \textit{QM} & 83.6 & 77.8 & 78.9\\
& \textit{Other} & 81.0 & 70.9 & 73.0\\
\end{tabular}
}
\caption{Derivation generation results by field of Physics. \textit{EM} is Electromagnetism, \textit{QM} is Quantum Mechanics, and \textit{Other} contains derivations from Classical and Statistical Mechanics.}
\label{tab:fields_scores}
\end{table}}

\noindent \textbf{Agreement between models on the relative difficulty of Physics domains.} The dataset is divided into the Electromagnetism, Quantum Mechanics, and Other subdomains (Tab.~\ref{tab:sizes}). According to the metrics, model derivations obtain highest scores on Electromagnetism and lowest on Other (Tab.~\ref{tab:fields_scores}). Notably, the quantum derivations are generally the most difficult and feature domain specific (braket) notation and relatively complex equational forms, so it is unexpected that models generate more coherent reasoning than in other subdomains, as the metrics suggest. A manual evaluation of GPT-4 derivations confirms these scores are misleading (Tab.~\ref{tab:gpt4_analysis}).

{\renewcommand{\arraystretch}{1.1}%
\begin{table}[h!]
	\centering
	\scalebox{0.9}{
		\begin{tabular}{@{}c|c|c @{}}
			\textbf{Physics} & \textbf{\# Derivations} & \textbf{Accuracy}\\
			\hline
            Electromagnetism & 76 & 88\\
            Quantum Mechanics & 74 & 69\\
            Other Physics & 29 & 83\\
            \hline
            All & 179 & 79\\
		\end{tabular}
		}
	\caption{Manual evaluation of 179 derivations generated by few-shot GPT-4.}
	\label{tab:gpt4_analysis}
\end{table}}

\begin{figure*}[h!]
    \centering
      % Adjust this to move the figure to the left
    \includegraphics[width=0.93\linewidth]{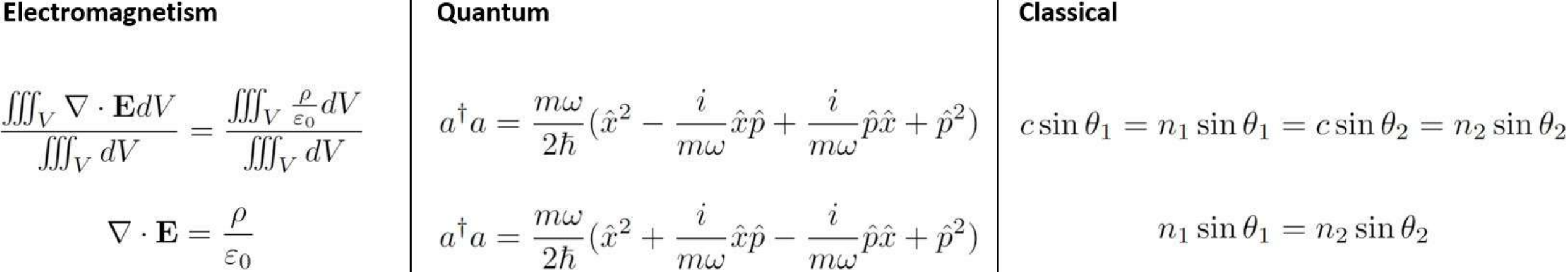}
    \caption{Excerpts from few-shot GPT-4 derivations that violate well-documented Physics.}
    \label{fig:Physics_errors}
\end{figure*}

\noindent To supplement previous analysis, we manually evaluate the coherence of the top scoring model's derivations in each subdomain (Tab.~\ref{tab:gpt4_analysis}), and describe instances where reasoning failures \textit{violate laws of Physics} without necessarily including mathematical errors (Fig.~\ref{fig:Physics_errors}). %We broadly consider a derivation to be ``coherent'' if it features no incorrect equations, does not skip many steps, or derives too many irrelevant or redundant equations. We do not expect output to perfectly match the ground truth, and we are lenient if a clear reasoning path can be identified. Many coherent derivations would certainly not obtain full marks. 
Notably, the \textit{\textbf{misuse of minus signs}} frequently contributes to incoherent reasoning, and \textit{\textbf{equations can be recalled incorrectly even if they are given in the prompt}}.

\paragraph{Use of Mathematics which violates Physics.} The coherence of a derivation does not depend on the correct application of Mathematics alone. Fundamental physical assumptions guide which mathematical steps are allowed, and in many cases, such as those in Fig.~\ref{fig:Physics_errors}, GPT-4 fails to understand this. For example, the \textbf{Electromagnetism} excerpt is only valid if the electric field $\mathbf{E}$ and charge density $\rho$ are constant throughout space. In this case, all integrals indeed cancel to give Gauss' law. As this law is true when quantities are spatially dependent, this reasoning is flawed. In the \textbf{Quantum} excerpt, both equations are true simultaneously only if $\hat{x}$ and $\hat{p}$ commute. Assuming each are respectively the quantum mechanical operators representing position and momentum, GPT-4 violates Heisenberg's uncertainty principle. The \textbf{Classical} excerpt is from a derivation that attempts to obtain Snell's law. GPT-4 asserts that light cruises across optical boundaries at 1 m/s without refraction. 

\subsection{Premise Removal}
\label{sec:premise_removal}

{\renewcommand{\arraystretch}{1}%
\begin{table*}[h!]
\centering
\scalebox{0.9}{
\begin{tabular}{@{}l|ccc|ccc|ccc @{}}
& \multicolumn{3}{c|}{\textbf{ROUGE}} & \multicolumn{3}{c|}{\textbf{BLEU}} & \multicolumn{3}{c}{\textbf{GLEU}}\\ 
& $S=0$ & $S=1$ & $S=2$ & $S=0$ & $S=1$ & $S=2$ & $S=0$ & $S=1$ & $S=2$ \\ \hline
MathT5-base & 69.5 & 64.1 & 56.4 & 58.5 & 55.1 & 49.0 & 59.9 & 56.7 & 51.3  \\
MathT5-large & 69.1 & 64.4 & 55.7 & 60.6 & 56.4 & 46.8 & 62.1 & 58.1 & 49.9 \\
GPT-3.5 & 77.7 & 75.8 & 60.2 & 68.7 & 65.6 & 44.4 & 70.5 & 68.4 & 51.5 \\
GPT-4 & \textbf{85.6} & \textbf{81.0} & \textbf{70.5} & \textbf{79.7} & \textbf{73.8} & \textbf{58.9} & \textbf{80.6} & \textbf{75.7} & \textbf{63.5} \\
\end{tabular}
}
\caption{Results from the Premise Removal perturbation analysis.}
\label{tab:perturbation_scores}
\end{table*}}

Following from Eq.~\ref{sec:equation}, we progressively remove premises from prompts and report corresponding scores in Tab.~\ref{tab:perturbation_scores}.  

\paragraph{The non-linear degradation in derivation quality reported by text generation metrics is supported by manual evaluation.} Across all metrics, the average performance degradation due to premise removal is non-linear with respect to the number of premises removed, and the score decrease from $(S = 0) \rightarrow (S = 1)$ is on average less than that of $(S = 1) \rightarrow (S = 2)$. However, this alone does not tell us the extent that removing premises \textit{leads to mathematical errors}, as it may be the case that models select alternative derivation paths which inherently lead to lower scores, despite the correct use of mathematics~\cite{meadows2023generating}. 

{\renewcommand{\arraystretch}{1}%
\begin{table}[h!]
	\centering
	\scalebox{1}{
		\begin{tabular}{@{}c|c|c|c @{}}
			 & $S = 0$ & $S = 1$ & $S = 2$\\
			\hline
            GPT-4 & 83 & 73 & 57\\
		\end{tabular}
		}
	\caption{GPT accuracy as $S$ premises are removed from the prompt.}
	\label{tab:gpt4_perturbed_analysis}
\end{table}}

To better assess the effect of premise removal, we manually evaluate 300 derivations for mathematical coherence. As with Tab.~\ref{tab:gpt4_analysis}, our evaluation is \textit{reference-free}~\cite{deutsch2022limitations,ke2022ctrleval,zhao2020limitations} with respect to the reasoning itself, although premises and goal equations are necessarily compared. Notably, we allow one or two missing steps given a coherent path can be traced through the derivation, and we are lenient if a derivation's underlying physics is not appropriately considered. The focus of the evaluation is the \textit{correct use of mathematics}, but sometimes the physical context requires specific mathematical behaviour (e.g., Fig.~\ref{fig:Physics_errors}).

To give helpful examples, in the supplementary material\footnote{\url{https://github.com/jmeadows17/transformers-for-physics/blob/main/Supplementary_Model_Derivations.pdf}}, the key concept in C.89 is to equate kinetic energy with work and rearrange for velocity, but rather than introduce kinetic energy as a premise, GPT \textit{reverse-engineers} the derivation from the goal equation, which we mark as correct. The derivations A.54 and B.54 would be marked as correct if they did not skip the premise $z = \braket{f}{g}$, but the reasoning as it stands would allow $z \geq \braket{f}{g}$, which is false.

Overall, the manual evaluation gives a \textit{liberal estimate} of model accuracy on the dataset in Tab.~\ref{tab:gpt4_perturbed_analysis}. We would expect Physics professionals to give lower scores, particularly for $S = 2$. The non-linear degradation in accuracy due to the gradual removal of premises aligns closely with scores obtained from the automatic metrics (notably BLEU). 

\paragraph{Substitution errors.} A fundamental component of mathematical reasoning is the substitution of equivalent terms. A significant proportion of errors involve substitution. 

\paragraph{Language models derive equations by reverse-engineering.} Working backwards from the result, this approach is characterised by a lack of understanding of the problem. Due to the prevalence of this reasoning in the outputs, this is perhaps the core mechanism behind how language models derive given equations. An example is given below:
\vspace{2pt}
\begin{addmargin}[1em]{1em}% 1em left, 2em right
$\sigma^2_p = \int^{\infty}_{-\infty}p^2|\phi(p)|^2 dp$ \;\;\; (initial premise)\newline \newline $\bra{g} = \int^{\infty}_{-\infty}p^2|\phi(p)|^2 dp$ \newline \newline $\ket{g} = 1$ \newline \newline $\sigma_p^2 = \braket{g}{g}$ \;\;\;\;\;\;\;\;\;\;\;\;\;\;\;\;\;\;\;\; (goal equation)
\end{addmargin}
\vspace{10pt}

\noindent \newline GPT-4 has clearly recognised that it must define terms for $\bra{g}$ and $\ket{g}$, but a fundamental lack of understanding of Dirac notation (from quantum mechanics) has resulted in defining them as scalar quantities instead of vectors.

\section{Conclusion}

We explore the mathematical ability of LMs in the context of a derivation generation task using a novel dataset of equation derivations and prompt interventions. The difference between reasoning steps is only one or two operations, with the intent of emulating granular mathematical workings and exploring how models perform detailed calculations involving a variety of Physics notations. 

We apply premise removal interventions on prompts to reveal a non-linear relationship between the perturbation strength and average derivation quality, as reported by manual and automatic scores. We find that models (particularly GPT-4) derive goal equations from premises through reverse-engineering intermediate steps without appropriate consideration of basic underlying Physics. This becomes increasingly apparent as we progressively remove premises from few-shot prompts. 

Furthermore, many algebraic errors arise from attempts at \textit{substitution}. While it is challenging to build synthetic datasets that reflect all aspects of mathematical reasoning~\cite{toshniwal2024openmathinstruct}, it is more manageable to focus on the application of individual operations (such as substitution), and use symbolic engines to apply them in the context of vast vocabularies of symbols and notations representative of target subdomains, which could potentially bolster models' out-of-distribution mathematical abilities. Our use of synthetic in-context examples (involving substitution) improves models' evaluation scores.   

Although exploration of the synergy between language models and symbolic engines is underway~\cite{yang2024leandojo,davis2024mathematics}, Physics derivations rely on physical behaviour which is implicitly assumed (\textit{e.g.,} commutative properties of quantum operators), yet is ignored by LMs. Such assumptions govern the reasoning paths allowed in derivations, and are hence independent of the ability to perform fine-grained algebraic manipulation. This is effectively a premise/operation selection problem, and integrating symbolic solvers does not inherently improve this limitation~\cite{liu2024augmenting}. We find contemporary language models severely lacking in this regard, and suggest research efforts should be directed towards LM-based querying of appropriate knowledge bases during inference alongside solver integration~\cite{trinh2024solving}.

\section{Limitations}

We have not extensively optimised prompts for model performance. However, our few-shot approach mirrors related work~\cite{azerbayev2023llemma,meadows2023generating}. Our focus on fine-grained mathematical transformations between equations excludes natural language (\textit{e.g.,} Fig.~\ref{fig:proof_comp}) as a design choice. Text generation metrics (\textit{e.g.,} ROUGE) fail to account for mathematical errors in individual derivations, however we contrast this with a human evaluation of 300 derivations in the case of GPT, which supports our conclusions. Due to the small number of human annotators responsible for curating the dataset and manually evaluating model outputs, both elements of this research may vary with the number of annotators or their experience in Physics. Our focus in this regard is a self-consistent set of derivations and assessment criteria agreed upon by the annotators involved. 

\paragraph{Overall ethical impact.} This work explores a systematic way to elicit the mathematical/symbolic inference properties of Transformer-based models in a mathematical language processing task. As such, it contributes in the direction of a critique of the reasoning capabilities and the biases of these models, particularly in the Physics domain.

\bibliography{anthology,custom}
\bibliographystyle{acl_natbib}

\appendix

\section{Further details on models, metrics, and prompting}
\label{app:models_and_metrics}

\paragraph{Models.} T5~\cite{raffel2020exploring} is an encoder-decoder transformer where all pre-training objectives are formulated as text generation (and therefore do not require different loss functions). FLAN-T5~\cite{chung2022scaling} is T5 fine-tuned on instructions, and outperforms T5 in a variety of tasks. The GPT models~\cite{brown2020language} are decoder-only transformer-based models trained on large-scale natural (and mathematical) language corpora. We evaluate 8 models on derivation generation: the base and large variants of T5 and FLAN-T5, GPT-3.5, GPT-4, MathT5-base, and MathT5-large~\cite{meadows2023generating}. MathT5-large is a version of FLAN-T5-large fine-tuned for 25 epochs on 15K (LaTeX) synthetic mathematical derivations (containing 4 - 10 equations), that were generated using a symbolic engine. It outperforms the few-shot performance of GPT-4 and GPT-3.5 on derivation generation in ROUGE, BLEU, BLEURT, and GLEU scores, and shows some generalisation capabilities. It was trained on 155 Physics-related symbols, but struggles with out-of-vocabulary symbols. MathT5-base is the equivalent, but uses T5-base as the initialised model before fine-tuning. Instantiated few-shot prompts are fed to the instruction-based models (in this case GPT) models through the OpenAI API\footnote{\url{https://platform.openai.com/overview}}, with temperature set to 0 to minimise non-deterministic effects.

\paragraph{Computation and hyperparameters.} Hyperparameters used in T5 models are the default hyperparameters of MathT5 defined in the MathT5.py script on the Hugging Face website. There was no training involved in the experiments of this paper, and models were evaluated on a single GTX 1070 for up to a week in total.  

\paragraph{Zero-shot prompts.} We prefix the following sentence to the prompt template discussed in the main paper: 

\noindent \textit{``Derive the final equation using the premise equations from the following prompt (denoted by "Prompt:"). Give only the equations involved in the derivation. Do not include any text other than equations each separated by "and". Prompt: ''}.

\paragraph{Few-shot prompts.} For each initial prompt, such as the example template in the main paper, a set of 5 example templates (and their derivations) are randomly selected from the set of \textit{synthetic derivations}. We select in-context examples by filtering only those containing \textit{more than one premise}, and with no given intermediate equations, to better match the Physics prompts. The examples are then fit into the few-shot prompt below:

\textit{The following examples consist of a prompt (denoted by Prompt:) and a mathematical derivation (denoted by Derivation:). Each derivation contains LaTeX equations separated by "and".} \newline \noindent The training prompts are appended after this description, then continues: \newline \indent \textit{Now given the following prompt, generate the derivation. Ensure equations are split by the word "and".} \newline \noindent The evaluation prompt is inserted here, prefixed by \textit{``Prompt:''}.

\paragraph{Metrics.} ROUGE, BLEU, and GLEU are all metrics used to evaluate the quality of text generated by machine translation or other natural language processing tasks, but they differ in their approaches and specific applications.

ROUGE (Recall-Oriented Understudy for Gisting Evaluation)~\cite{lin2004rouge} focuses on the overlap of n-grams, word sequences, and word pairings between the generated text and reference texts, emphasising \textit{recall}. It is widely used in summarisation evaluation and other tasks where capturing the essence of the reference material is critical. BLEU (Bilingual Evaluation Understudy)~\cite{papineni2002bleu} measures the \textit{precision} of n-gram matches between the output and reference texts, adjusted by a brevity penalty to discourage overly short translations. BLEU is predominantly used in machine translation to assess the closeness of the translation to human-produced texts. 

GLEU (Generalized Language Understanding Evaluation)~\cite{mutton2007gleu} is similar to BLEU in its use of n-gram overlap but was specifically designed for evaluating grammatical error corrections. GLEU includes modifications to accommodate the nuances of grammar correction by considering both the presence of corrected n-grams and penalising uncorrected errors, without the need for tuning across different numbers of reference texts. While ROUGE emphasises capturing the essence of the text through recall, BLEU focuses on precision, and GLEU targets the specific domain of grammatical correctness. The latter is used in related work~\cite{welleck2022naturalprover}. 

\section{Electromagnetism}
\label{app:electromagnetism}

\subsection{Gauss' law: Derivation from Coulomb's law}

\begin{equation}
%PREM
    \mathbf{E}(\mathbf{r}) = \frac{1}{4\pi\varepsilon_0} \int \frac{\rho(\mathbf{s})(\mathbf{r} - \mathbf{s})}{|\mathbf{r} - \mathbf{s}|^3}d^3\mathbf{s}
\end{equation}

\begin{equation}
    \nabla \cdot \mathbf{E}(\mathbf{r}) = \nabla \cdot \frac{1}{4\pi\varepsilon_0} \int \frac{\rho(\mathbf{s})(\mathbf{r} - \mathbf{s})}{|\mathbf{r} - \mathbf{s}|^3}d^3\mathbf{s}
\end{equation}

\begin{equation}
    \nabla \cdot \mathbf{E}(\mathbf{r}) = \frac{1}{4\pi\varepsilon_0} \int \rho(\mathbf{s}) \nabla \cdot \frac{(\mathbf{r} - \mathbf{s})}{|\mathbf{r} - \mathbf{s}|^3}d^3\mathbf{s}
\end{equation}

\begin{equation}
%PREM
    \nabla \cdot \frac{(\mathbf{r} - \mathbf{s})}{|\mathbf{r} - \mathbf{s}|^3} = 4 \pi \delta (\mathbf{r} - \mathbf{s})
\end{equation}

\begin{equation}
    \nabla \cdot \mathbf{E}(\mathbf{r}) = \frac{1}{4\pi\varepsilon_0} \int \rho(\mathbf{s}) 4 \pi \delta (\mathbf{r} - \mathbf{s})d^3\mathbf{s}
\end{equation}

\subsection{Uniqueness theorem for Poisson's equation 2}

\begin{equation}
%PREM
    \nabla \cdot (\phi\nabla\phi) = (\nabla\phi)^2 + \phi\nabla^2\phi
\end{equation}

\begin{equation}
%PREM
    \nabla^2\phi = 0
\end{equation}

\begin{equation}
    \nabla \cdot (\phi\nabla\phi) = (\nabla\phi)^2
\end{equation}

\begin{equation}
    \int_V \nabla \cdot (\phi\nabla\phi) dV = \int_V (\nabla\phi)^2 dV
\end{equation}

\begin{equation}
%PREM
    \int_V \nabla \cdot (\phi\nabla\phi) dV = \int_S \phi\nabla\phi \cdot d\mathbf{S}
\end{equation}

\begin{equation}
    \int_S \phi\nabla\phi \cdot d\mathbf{S} = \int_V (\nabla\phi)^2 dV
\end{equation}

\subsection{Lorentz force: Derivation of Lorentz force from classical Lagrangian (LHS)}

\begin{equation}
%PREM
    L = \frac{m}{2}(\dot{x}^2 + \dot{y}^2 + \dot{z}^2) + q(\dot{x}A_x + \dot{y}A_y + \dot{z}A_z) - q\phi
\end{equation}

\begin{equation}
    \frac{\partial L}{\partial \dot{x}} = \frac{\partial}{\partial \dot{x}}(\frac{m}{2}(\dot{x}^2 + \dot{y}^2 + \dot{z}^2) + q(\dot{x}A_x + \dot{y}A_y + \dot{z}A_z) - q\phi)
\end{equation}

\begin{multline}
    \frac{\partial L}{\partial \dot{x}} = \frac{m}{2}\frac{\partial}{\partial \dot{x}}(\dot{x}^2 + \dot{y}^2 + \dot{z}^2)\\ + q\frac{\partial}{\partial \dot{x}}(\dot{x}A_x + \dot{y}A_y + \dot{z}A_z) - q\frac{\partial}{\partial \dot{x}}\phi
\end{multline}

\begin{equation}
%PREM
    \frac{\partial}{\partial \dot{x}}\phi = 0
\end{equation}

\begin{multline}
    \frac{\partial L}{\partial \dot{x}} = \frac{m}{2}\frac{\partial}{\partial \dot{x}}(\dot{x}^2 + \dot{y}^2 + \dot{z}^2)\\ + q\frac{\partial}{\partial \dot{x}}(\dot{x}A_x + \dot{y}A_y + \dot{z}A_z)
\end{multline}

\subsection{Ampere's circuital law: Proof of equivalence 2}

\begin{equation}
%PREM
    \nabla\times \mathbf{H} = \mathbf{J}_f + \frac{\partial \mathbf{D}}{\partial t}
\end{equation}

\begin{equation}
%PREM
    \nabla \times \frac{1}{\mu_0}\mathbf{B} = \nabla \times \mathbf{H} + \mathbf{J}_M
\end{equation}

\begin{equation}
    \nabla \times \frac{1}{\mu_0}\mathbf{B} = \mathbf{J}_f + \frac{\partial \mathbf{D}}{\partial t} + \mathbf{J}_M
\end{equation}

\begin{equation}
%PREM
    \mathbf{D} = \varepsilon_0\mathbf{E} + \mathbf{P}
\end{equation}

\begin{equation}
    \frac{\partial \mathbf{D}}{\partial t} = \frac{\partial}{\partial t}(\varepsilon_0\mathbf{E} + \mathbf{P})
\end{equation}

\begin{equation}
    \frac{\partial \mathbf{D}}{\partial t} = \frac{\partial}{\partial t}\varepsilon_0\mathbf{E} + \frac{\partial}{\partial t}\mathbf{P}
\end{equation}

\begin{equation}
    \nabla \times \frac{1}{\mu_0}\mathbf{B} = \mathbf{J}_f + \frac{\partial}{\partial t}\varepsilon_0\mathbf{E} + \frac{\partial}{\partial t}\mathbf{P} + \mathbf{J}_M
\end{equation}

\section{Quantum Mechanics}
\label{app:quantum}

\subsection{Uncertainty principle: Kennard inequality proof part 2.7}

\begin{equation}
%PREM
    g(x) = \frac{1}{2\pi\hbar}\int_{-\infty}^{\infty}p\cdot (\frac{\hbar}{ip} \int^{\infty}_{-\infty}\frac{d\psi(\chi)}{d\chi}e^{\frac{-ip\chi}{\hbar}}d\chi)\cdot e^{\frac{ipx}{\hbar}}dp
\end{equation}

\begin{equation}
    g(x) = \frac{1}{2\pi\hbar}\int_{-\infty}^{\infty}\frac{\hbar}{i} \int^{\infty}_{-\infty}\frac{d\psi(\chi)}{d\chi}e^{\frac{-ip\chi}{\hbar}}d\chi\cdot e^{\frac{ipx}{\hbar}}dp
\end{equation}

\begin{equation}
    g(x) = \frac{1}{2\pi i}\int_{-\infty}^{\infty} \int^{\infty}_{-\infty}\frac{d\psi(\chi)}{d\chi}e^{\frac{-ip\chi}{\hbar}}d\chi\cdot e^{\frac{ipx}{\hbar}}dp
\end{equation}

\begin{equation}
    e^{\frac{-ip\chi}{\hbar}}\cdot e^{\frac{ipx}{\hbar}} = e^{\frac{i}{\hbar}(x-\chi)p}
\end{equation}

\begin{equation}
    g(x) = \frac{1}{2\pi i}\int_{-\infty}^{\infty} \int^{\infty}_{-\infty}\frac{d\psi(\chi)}{d\chi}e^{\frac{i}{\hbar}(x-\chi)p}d\chi dp
\end{equation}

\subsection{Uncertainty principle: Kennard inequality proof part 3.3}

\begin{equation}
%PREM
    \sigma_p^2 = \int_{-\infty}^{\infty}p^2|\varphi(p)|^2dp
\end{equation}

\begin{equation}
%PREM
    |\tilde{g}(p)|^2 =  p^2|\varphi(p)|^2
\end{equation}

\begin{equation}
    \sigma_p^2 = \int_{-\infty}^{\infty}|\tilde{g}(p)|^2dp
\end{equation}

\begin{equation}
%PREM
    \int_{-\infty}^{\infty}|\tilde{g}(p)|^2dp = \int_{-\infty}^{\infty}|g(x)|^2dx
\end{equation}

\begin{equation}
    \sigma_p^2 = \int_{-\infty}^{\infty}|g(x)|^2dx
\end{equation}

\begin{equation}
%PREM
   \bra{g}\ket{g} = \int_{-\infty}^{\infty}|g(x)|^2dx
\end{equation}

\begin{equation}
    \sigma_p^2 = \bra{g}\ket{g}
\end{equation}

\subsection{Expectation value: integral expression 3}

\begin{multline}
%PREM
     \expval{\hat{X}}_{\Psi} = \int \int \braket{x}{\Psi}^{\dagger}x'\delta(x-x') \\ \braket{x'}{\Psi}dx dx'
\end{multline}

\begin{multline}
%PREM
     \int \braket{x}{\Psi}^{\dagger}x'\delta(x-x') \braket{x'}{\Psi} dx' \\ = \braket{x}{\Psi}^{\dagger}x \braket{x}{\Psi}
\end{multline}

\begin{equation}
    \expval{\hat{X}}_{\Psi} = \int \braket{x}{\Psi}^{\dagger}x\braket{x}{\Psi} dx
\end{equation}

\begin{equation}
%PREM
    \braket{x}{\Psi} = \Psi(x)
\end{equation}

\begin{equation}
    \expval{\hat{X}}_{\Psi} = \int \Psi^{\dagger}(x) x \Psi(x) dx
\end{equation}

\begin{equation}
%PREM
    \Psi^{\dagger}(x) \Psi(x) = |\Psi(x)|^2
\end{equation}

\begin{equation}
    \expval{\hat{X}}_{\Psi} = \int x |\Psi(x)|^2 dx
\end{equation}

\subsection{Hellmann–Feynman theorem 2}

\begin{equation}
%PREM
    \frac{d E_\lambda}{d\lambda} = \frac{d}{d\lambda} \bra{\Psi_\lambda} \hat{H}_\lambda\ket{\Psi_\lambda}
\end{equation}

\begin{multline}
%PREM
    \frac{d}{d\lambda} \bra{\Psi_\lambda} \hat{H}_\lambda\ket{\Psi_\lambda} = \bra{\frac{d \Psi_\lambda}{d\lambda}} \hat{H}_\lambda\ket{\Psi_\lambda} \\ + \bra{\Psi_\lambda} \frac{d\hat{H}_\lambda}{d\lambda} \ket{\Psi_\lambda}\\ + \bra{\Psi_\lambda} \hat{H}_\lambda\ket{\frac{d \Psi_\lambda}{d\lambda}}
\end{multline}

\begin{multline}
    \frac{d E_\lambda}{d\lambda} = \bra{\frac{d \Psi_\lambda}{d\lambda}} \hat{H}_\lambda\ket{\Psi_\lambda} \\ + \bra{\Psi_\lambda} \frac{d\hat{H}_\lambda}{d\lambda} \ket{\Psi_\lambda}\\ + \bra{\Psi_\lambda} \hat{H}_\lambda\ket{\frac{d \Psi_\lambda}{d\lambda}}
\end{multline}

\begin{equation}
%PREM
    \bra{\Psi_\lambda}\hat{H}_\lambda = \bra{\Psi_\lambda}E_\lambda
\end{equation}

\begin{equation}
%PREM
    \hat{H}_\lambda\ket{\Psi_\lambda} = E_\lambda\ket{\Psi_\lambda}
\end{equation}

\begin{multline}
    \frac{d E_\lambda}{d\lambda} = E_\lambda \bra{\frac{d \Psi_\lambda}{d\lambda}}\ket{\Psi_\lambda} + \bra{\Psi_\lambda} \frac{d\hat{H}_\lambda}{d\lambda} \ket{\Psi_\lambda} \\ +  E_\lambda \bra{\Psi_\lambda}\ket{\frac{d \Psi_\lambda}{d\lambda}}
\end{multline}

\section{Other}
\label{app:other}

\subsection{Euler-Lagrange equation: Derivation 2}

\begin{equation}
%PREM
    \frac{dJ_{\varepsilon}}{d\varepsilon} = \int_{a}^{b}\frac{dL_{\varepsilon}}{d\varepsilon}dx
\end{equation}

\begin{equation}
%PREM
    \frac{dL_{\varepsilon}}{d\varepsilon} = \frac{\partial L_{\varepsilon}}{\partial g_{\varepsilon}}\eta(x) + \frac{\partial L_{\varepsilon}}{\partial g_{\varepsilon}'} \eta '(x)
\end{equation}

\begin{equation}
    \frac{dJ_{\varepsilon}}{d\varepsilon} = \int_{a}^{b}(\frac{\partial L_{\varepsilon}}{\partial g_{\varepsilon}}\eta(x) + \frac{\partial L_{\varepsilon}}{\partial g_{\varepsilon}'} \eta '(x))dx
\end{equation}

\begin{equation}
    \frac{dJ_{\varepsilon}}{d\varepsilon} = \int_{a}^{b}\frac{\partial L_{\varepsilon}}{\partial g_{\varepsilon}}\eta(x)dx + \int_{a}^{b}\frac{\partial L_{\varepsilon}}{\partial g_{\varepsilon}'} \eta '(x)dx
\end{equation}

\begin{equation}
    \eval{\frac{dJ_{\varepsilon}}{d\varepsilon}}_{\varepsilon = 0} = \eval{(\int_{a}^{b}\frac{\partial L_{\varepsilon}}{\partial g_{\varepsilon}}\eta(x)dx + \int_{a}^{b}\frac{\partial L_{\varepsilon}}{\partial g_{\varepsilon}'} \eta '(x)dx)}_{\varepsilon = 0}
\end{equation}

\begin{equation}
    \eval{\frac{dJ_{\varepsilon}}{d\varepsilon}}_{\varepsilon = 0} = \eval{(\int_{a}^{b}\frac{\partial L_{\varepsilon}}{\partial g_{\varepsilon}}\eta(x)dx)}_{\varepsilon = 0} + \eval{(\int_{a}^{b}\frac{\partial L_{\varepsilon}}{\partial g_{\varepsilon}'} \eta '(x)dx)}_{\varepsilon = 0}
\end{equation}

\subsection{Snell's law: from Fermat's principle}

\begin{equation}
%PREM
    T = \frac{(x^2 + a^2)^\frac{1}{2}}{v_1} + \frac{(b^2 + (l-x)^2)^\frac{1}{2}}{v_2}
\end{equation}

\begin{equation}
    T = \frac{(x^2 + a^2)^\frac{1}{2}}{v_1} + \frac{(b^2 + l^2 -2lx + x^2)^\frac{1}{2}}{v_2}
\end{equation}

\begin{equation}
    \frac{dT}{dx} = \frac{d}{dx} ( \frac{(x^2 + a^2)^\frac{1}{2}}{v_1} + \frac{(b^2 + l^2 -2lx + x^2)^\frac{1}{2}}{v_2})
\end{equation}

\begin{equation}
    \frac{dT}{dx} = \frac{d}{dx} ( \frac{(x^2 + a^2)^\frac{1}{2}}{v_1}) + \frac{d}{dx} (\frac{(b^2 + l^2 -2lx + x^2)^\frac{1}{2}}{v_2})
\end{equation}

\begin{equation}
%PREM
    \frac{d}{dx} ( \frac{(x^2 + a^2)^\frac{1}{2}}{v_1}) = \frac{x}{v_1 (x^2 + a^2)^\frac{1}{2}}
\end{equation}

\begin{equation}
%PREM
    \frac{d}{dx} (\frac{(b^2 + l^2 -2lx + x^2)^\frac{1}{2}}{v_2}) = \frac{x-l}{v_2  ((x-l)^2 + b^2)^\frac{1}{2} }
\end{equation}

\begin{equation}
    \frac{dT}{dx} = \frac{x}{v_1 (x^2 + a^2)^\frac{1}{2}} + \frac{x-l}{v_2  ((x-l)^2 + b^2)^\frac{1}{2} }
\end{equation}

\subsection{Maxwell-Boltzmann: energy distribution 2}

\begin{equation}
%PREM
    |\mathbf{p}|^2 d|\mathbf{p}| = m (2mE)^{\frac{1}{2}} dE
\end{equation}

\begin{equation}
%PREM
    d^3\mathbf{p} = 4 \pi |\mathbf{p}|^2 d|\mathbf{p}|
\end{equation}

\begin{equation}
    d^3\mathbf{p} = 4\pi m (2mE)^  {\frac{1}{2}} dE
\end{equation}

\begin{equation}
%PREM
    f_{\mathbf{p}}(\mathbf{p}) = (2\pi mkT)^{-\frac{3}{2}} e^{-\frac{\mathbf{p}^2}{2mkT}}
\end{equation}

\begin{equation}
    f_{\mathbf{p}}(\mathbf{p}) d^3\mathbf{p} = (2\pi mkT)^{-\frac{3}{2}} e^{-\frac{\mathbf{p}^2}{2mkT}} d^3\mathbf{p}
\end{equation}

\begin{equation}
    f_\mathbf{p}(\mathbf{p}) d^3\mathbf{p} = (2\pi mkT)^{-\frac{3}{2}} e^{-\frac{\mathbf{p}^2}{2mkT}} 4\pi m (2mE)^{\frac{1}{2}} dE
\end{equation}

\begin{equation}
%PREM
    f_E(E)dE = f_{\mathbf{p}}(\mathbf{p})d^3\mathbf{p}
\end{equation}

\begin{equation}
    f_E(E)dE = (2\pi mkT)^{-\frac{3}{2}} e^{-\frac{\mathbf{p}^2}{2mkT}} 4\pi m (2mE)^{\frac{1}{2}} dE
\end{equation}

\begin{equation}
    f_E(E) = 2 (\frac{E}{\pi})^{\frac{1}{2}} (\frac{1}{kT})^{\frac{3}{2}} e^{-(\frac{E}{kT})}
\end{equation}

\subsection{Wave equation: stress pulse in a bar}

\begin{multline}
%PREM
    \frac{\partial^2}{\partial t^2}u(x+h,t) = \frac{KL^2}{Mh^2} \big( u(x+2h,t) - 2u(x+h,t) \\ + u(x,t) \big)
\end{multline}

\begin{multline}
    \lim_{h\to 0} \frac{\partial^2 u(x+h,t)}{\partial t^2} = 
    \lim_{h\to 0}\frac{KL^2}{Mh^2} \big( u(x+2h,t) \\ - 2u(x+h,t) + u(x,t) \big)
\end{multline}

\begin{multline}
%PREM
    \lim_{h\to 0} \frac{u(x+2h,t) - 2u(x+h,t) + u(x,t)}{h^2} = \\ \frac{\partial^2 u(x,t)}{\partial x^2} 
\end{multline}

\begin{multline}
    \lim_{h\to 0} \frac{\partial^2 u(x+h,t)}{\partial t^2} = \frac{KL^2}{M} \frac{\partial^2 u(x,t)}{\partial x^2} 
\end{multline}

\begin{equation}
    \lim_{h\to 0} \frac{\partial^2 u(x+h,t)}{\partial t^2} = \frac{\partial^2 u(x,t)}{\partial t^2}
\end{equation}

\begin{equation}
    \frac{\partial^2 u(x,t)}{\partial t^2} =  \frac{KL^2}{M} \frac{\partial^2 u(x,t)}{\partial x^2} 
\end{equation}

\end{document}